\documentclass[conference]{IEEEtran}

\makeatletter

\def\ps@IEEEtitlepagestyle{%
  \def\@oddfoot{\mycopyrightnotice}%
  \def\@evenfoot{}%
}
\def\mycopyrightnotice{%
  \gdef\mycopyrightnotice{}
}

\usepackage{blindtext}
\usepackage{eso-pic}
\IEEEoverridecommandlockouts
\usepackage{cite}
\usepackage{amsmath,amssymb,amsfonts}
\usepackage{algorithmic}

\usepackage{graphicx}
\usepackage{textcomp}
\usepackage{xcolor}
\def\BibTeX{{\rm B\kern-.05em{\sc i\kern-.025em b}\kern-.08em
    T\kern-.1667em\lower.7ex\hbox{E}\kern-.125emX}}
    
\usepackage{eso-pic}

 \usepackage[ruled, vlined, linesnumbered]{algorithm2e}
\makeatletter
\newenvironment{tablehere}
{\def\@captype{table}}
{
	
}
\newenvironment{figurehere}
{\def\@captype{figure}}
{}
\makeatother
\begin{document}
\title{\vspace*{1cm} Evolutionary Extreme Learning Machine of ab-initio Energy Landscapes for Crystal Structure Prediction using Manta Ray Optimization with L\'{e}vy Flight\\
}

\author{\IEEEauthorblockN{Adrian Rubio-Solis}
\IEEEauthorblockA{\textit{Hamlyn Centre for Robotic Surgery, Imperial College London (ICL)} \\
\textit{London SW7 2AZ, UK} United Kingdom \\
arubioso@ic.ac.uk}
}

\maketitle
\begin{abstract}
The Manta Ray Foraging Optimization algorithm (MRFO) has proven to be a powerful heuristic strategy in the optimal solution of a large number of engineering problems. In this paper, an improvement of MRFO with L\'{e}vy Flight is suggested for the training of extreme learning machines (ELMs) whose basic model is a Single Layer Feedforward Network (SLFN). The proposed methodology that we called Evolutionary EELM-MRFO-LF for short is implemented to the prediction of unrelaxed and relaxed formation energy compounds relative to ground state crystal structure of pure components in binary systems. EELM-MRFO-LF follows the learning procedure of traditional Evolutionary ELMs in which first MRFO with LF is used to select the input weights and
Moore–Penrose (MP) generalized inverse is applied to analytically determine the output weights. L\'{e}vy Flight trajectory is implemented for increasing the diversity of the population of ELMs against premature convergence and the ability of avoiding getting trapped in a local optima. The performance of the suggested EELM-MRFO-LF is compared with other well-known nature-inspired algorithms under similar conditions.
\newline
\end{abstract}



\section{Introduction}
Extreme Learning Machine (ELM) was originally introduced as a new class of off-line learning methodology for the training of Single-hidden-Layer Feedforward Neural Networks (SLFNs), much faster and suitable to provide improved generalisation capabilities than traditional gradient descent learning approaches \cite{intro_1, intro_4,intro_5,intro_6} . ELM randomly selects the input weights and hidden biases of a SLFN and analytically determines its output weights through simple generalized inverse operation of the hidden layer output matrix. Moreover, ELM overcomes some drawbacks that face traditional gradient descent approaches such as stopping criteria, learning rate, learning epochs, and local minima \cite{intro_2, intro_4}.  However, it is also found that ELM usually tends to require more hidden neurons than traditional gradient-based learning algorithms as well as sometimes resulting in ill-condition problems due to the random selection of input weights and hidden biases \cite{intro_6_v1, intro_7}. To address this problem, numerous studies in the literature have proposed metaheuristic-based approaches as a mechanism to select input weights and biases and Moore–Penrose (MP) generalized inverse to analytically determine the output weights\cite{intro_4,intro_5,intro_6, intro_7_v1,intro_7_v4,intro_7_v2,intro_7_v3}. For example in \cite{intro_6_v1}, an evolutionary ELM with linear hidden neurons and using particle swarm optimization (PSO) demonstrated that ELM approach is able to achieve good generalization performance with much
more compact networks. PSO was applied to optimize the input weights and hidden biases of SLFNs to address prediction tasks, with boundary conditions primarily incorporated into the PSO framework while enhancing the performance of the ELM. In \cite{intro_7_v5}, authors proposed an enhanced ELM method to the modelling of monthly river flow from hydro-climatic data using reptile search optimization (RSA). The proposed method was also compared to a number of popular metaheuristics, including salp swarm algorithm (SSA) and equilibrium optimizer (EO). Results demonstrated an improved model accuracy of traditional ELM in the prediction of river flow using popular accuracy metrics such as Root-Mean-Square-Error (RMSE), Mean-Absolute-Error (MAE), Nash-Sutcliffe Efficiency (NSE) and Determination Coeffcient ($R^2$). Metaheuristic-based selection of input weights and biases improves the generalization performance of ELM and mitigates convergence to local minima.

MRFO has demonstrated its capability to solve a wide range of real world problems \cite{intro_7_v6}. It is also known, MRFO suffers from slow convergence precision and it is easily trapped in a local optimal \cite{qu2024improved}. In \cite{qu2024improved}, To overcome this, authors implemented Tent chaotic mapping, a bidirectional search and L\'{e}vy Flight. While Tent chaotic mapping distributes more uniformly the initial solutions, bidirectional search expands the search area. From results presented \cite{qu2024improved}, incorporating L\'{e}vy Flight clearly contributed the algorithm’s ability to escape from
local optimal. 

In this paper, a new method combining  Extreme Learning Machine and Manta Ray Foraging Optimization with L\'{e}vy Flight (MRFO-LF) and called EELM-MRF-LF for short is suggested for the prediction of formation energies in crystal structures relative to the pure components across the entire compositional and configurational space of binary systems. As described in \cite{b2}, predicting the crystal structure of a material given its constituent elements plays a crucial role to determine the total energy of the structure. Especially, predicting the atomic configuration that minimises the total energy usually results computational intensive using traditional techniques such as Density Functional Theory (DFT) \cite{b2, b3}. Machine learning has become an attractive computational tool not only to accelerate the structure search by recognising energy configurations with a much lower cost than DFT, but also to accurately predict unseen data \cite{b2}. The proposed EELM-MRFO-LF extends the work presented in \cite{b2}, by introducing a new machine learning strategy that provides an improved prediction accuracy and reduces the number of outliers. EELM-MRFO-LF is applied to iteratively determine the optimal parameters of SLFNs. A data set of 14,000 Li-Ge structures was employed.  As indicated in \cite{intro_7_v6} and \cite{intro_7_v7}, MRFO usually suffers from slow convergence precision and is easily trapped in a local optima. To address this limitation, LF was incorporated into the MRFO algorithm to enhance its ability to escape local optima. \cite{ intro_7_v7}.

In this work, EELM-MRFO-LF adopts LF to increase the diversity of search agents during somersault  foraging ensuring that the global search achieves a higher exploration performance and avoids getting trapped in a local optima. To evaluate the performance of EELM-MRFO-LF  with respect to other popular metaheuristics, techniques such as Whale Optimization Algorithm (WOA) \cite{ELM_WOA}, continuous Genetic Algorithm (GA) \cite{ELM_AG}, Particle swarm optimization (PSO) \cite{intro_2}  and traditional ELM \cite{intro_2} were also implemented.

The rest of this paper is organised as follows. In section II, background material is revisited and the proposed EELM-MRFO-LF is also described. In section III, experiments and discussions are presented, while in section IV conclusions are drawn.  
\section{METHODS}

\subsection{Dataset}

In this paper, the data set generated in \cite{b2} is used to test the proposed evolutionary ELM using Manta Ray Foraging Optimization with L\'{e}vy Flight (EELM-MRFO-LF). Such data consists of approximately 14,000 Li-Ge structures generated by a Genetic Algorithm for structure prediction called GASP for short \cite{b3}. As described in \cite{b2}, GASP generates an initial population of structures and employs DFT to evaluate their associated energy. Relevant energy for structure predictions as a function of composition in the formation energy relative to the crystal of pure components, i.e.:
\begin{equation}
    E_f = E_{tot} - X_{Li}E_{Li} - X_{Ge}E_{Ge},  
\end{equation}
in which, $E_{tot}$ is the total energy per atom of the Li-Ge crystal structure, $X_{Li}$ and $X_{Ge}$ are the molar fractions of $Li$ and $Ge$ in the structure. $E_{Li}$ and $E_{Ge}$ are the energy per atom of pure Li and Ge. For comparison reasons, this paper the term 'unrelaxed' formation energy ($E_f^u$) is employed to denote the energy of unrelaxed structures, while the term relaxed formation energy ($E_f^r$) refers to the energy of the minimum-energy structure obtained upon relaxation. In our experiments, the data selected from structures with unrelaxed formation energies $E_f^U < 200meV/atom$ is employed,  resulting in a final data set of 14,168 crystal structures. 
\subsection{Data Representation}
The proposed EELM-MRFO-LF is applied to build a regression model of energy data based on a Multiple-Input-Single-Output SLFN. As suggested in \cite{b2}, to achieve this, a data representation of the input data must be first constructed. In this paper, data representation follows the strategy used in \cite{b2}, where a Partial Radial Function (RDF) $g_{AB}(r)$ is computed to capture the atomic distances $d_{ij}^{AB} =|\overrightarrow{r}_i^A  - \overrightarrow{r}_j^B|$ between atoms $i$ and $j$ of type A and B as follows:
\begin{equation}
    g_{AB}(r) = \frac{1}{N_A}\sum_{i=1}^{N_A} \sum_{j=1}^{\infty}  \frac{1}{r^p} exp \left[- \frac{(r - d_{ij}^{AB} )^2}{2 \sigma_g^2} \right] \Theta (d_c - d_{ij}^{AB})
\end{equation}
Where $N_A$ includes all atoms of type A within the unit cell, while the second sum consists of all atoms of type B up to a cutoff distance $d_c$. $\Theta (d_c - d_{ij}^{AB})$ is the Heaviside function, $1/r^p$ renormalizes the  partial RDF as a function of distance. $\sigma_g$ is the width of a Gaussian function used to adjust data distribution defined in this work as $\sigma_g = 0.2\mathring{A}$.   
\subsection{Extreme Learning Machine}\label{AA}
Extreme Learning Machine (ELM) is a learning paradigm originally developed to train single-hidden-layer feedforward networks (SLFNs), in which parameters in the hidden neurons are initialized randomly and the output weights are optimized using the Moore-Penrose pseudoinverse. Given a number of  $'P'$ distinct samples $D = (\textbf{x}_p, t  _p) $, with each $\textbf{x}_p$ being a N dimensional vector and $t_p$ as the target scalar output. Hence the goal of ELM is to find a relationship between $\textbf{x}_p$ and $\textbf{t}_p$. Standard SLFNs with $M$ hidden nodes and activation $h( \cdot )$ function can be mathematically modelled by \cite{ARS_1A}:
\begin{equation}
\sum_{k=1}^{M} \beta_k h_k( \textbf{w}_k; \textbf{x}_p ) = \textbf{h}( \textbf{w}_k; \textbf{x}_p )\pmb{\beta} = y_p, ~~1 \leq p \leq P
\end{equation}
in which $\textbf{h}(\textbf{w}_k; \textbf{x}_p)=[h_1(\textbf{w}_1; \textbf{x}_1), \ldots, h_M(\textbf{w}_M; \textbf{x}_M) ]$ is the hidden feature mapping, $\textbf{w}_k = [\textbf{w}_{1}, \ldots, \textbf{w}_{N}]^T$ is the weight vector a randomly generated parameter of the hidden layer connecting the $kth$ hidden node and the input nodes. The output weight $\beta_p = [\beta_{p1}, \ldots, \beta_{p\tilde{N}} ]^T$ is the weight vector connecting the $kth$ hidden node to the $nth$ output. A SLFN with $M$ hidden nodes and activation function $g(\textbf{x})$ can approximate $P$ samples with zero error means $\sum_{p=1}^M \parallel \textbf{y}_p -  \textbf{t}_p \parallel$. Thus, a matrix representation of Eq. (3) is \cite{ARS_1A}:
\begin{equation}
\textbf{H}
 =  \begin{pmatrix}
  h(\textbf{w}_1; \textbf{x}_1) & \cdots  & h(\textbf{w}_k; \textbf{x}_p) \\
  \vdots  & \vdots  & \vdots\\
  h(\textbf{w}_k; \textbf{x}_p) & \cdots  & h(\textbf{w}_k; \textbf{x}_p) \\
 \end{pmatrix}_{P \times M}
 \end{equation}

 Where \textbf{H} is the hidden matrix of an SLFN with respect to the inputs $\textbf{x}_p$. The target vector is defined by $\textbf{T} = [t_1, \ldots, t_P]$.   The minimum norm least-squares solution of the linear system $\textbf{H} \beta = \textbf{T}$ is unique and can be achieved by calculating the Moore-Penrose pseudo-inverse $H^{\dagger}$ as:
\begin{equation}
\hat{\pmb{\beta}} = \textbf{H}^{\dagger} \textbf{T}
\end{equation}
In which, $\textbf{H}^{\dagger}$ can be calculated using the orthogonal projection method: $\textbf{H}^{\dagger}= (\textbf{H}^T \textbf{H})^{-1}\textbf{H}^T$ when $\textbf{H}^T \textbf{H}$ is nonsingular, or $\textbf{H}^{\dagger}= \textbf{H}(\textbf{H} \textbf{H}^T)^{-1}\textbf{H}^T$ when $\textbf{H} \textbf{H}^T$ is nonsingular. A penalty term can be added to the diagonal of $\textbf{H}^T \textbf{H}$ or $\textbf{H} \textbf{H}^T$ for regularization purpose. However, the optimum value of this penalty is subjected the minimization of the validation error. 
%

\subsection{Manta Ray Foraging Optimization}
The original MRFO algorithm was inspired on intelligent foraging behaviours of manta rays. As detailed in Algorithm 1, the mathematical model of this algorithm consists of three main foraging operators used to mimic manta ray's hunt for plankton and food, namely: a) chain foraging, b) cyclone foraging and c) somersault foraging \cite{MRFO_1}. 

\subsubsection{Chain foraging}
In this behaviour, manta rays search for plankton (food) swimming towards it once they determine their position. The best position is that one containing the higher plankton concentration. Manta rays queue together head-to-tail forming a foraging chain. Each manta's move is both towards the food and to the individual in front of it. That is, in every iteration, each manta is updated by the best solution found so far and the solution in front of it. Chain foraging is implemented as described in Eq. (7) in Algorithm 1, in which, the term $\beta = 2 \cdot r \cdot \sqrt{|log(r)|}$, where $\textbf{x}_i(t) = (x_i^1(t), \ldots, x_i^D(t))$ is the vector position of the $ith$ individual at time t and $D$ its dimension. The term $r$ is a random vector within the range [0, 1], $\alpha$ is a weight coefficient, $\textbf{x}_{best}$ is the plankton location with high concentration. As described in \textbf{Algorithm 1}, $\textbf{x}_{i-1}(t)$ of the $(i-1)th$ current individual position and the best position $x_{best}$ of the food \cite{MRFO_1}. 

\subsubsection{Cyclone foraging}
In this stage, when the school of mantas recognise a patch of plankton in deep waters, the school will form a long foraging chain and swim towards the food by a spiral. In this behaviour, the school follows a spiral foraging while moving towards the food where each manta ray swims towards the one in front of it. This means, each manta swarms in line developing a spiral perform foraging \cite{MRFO_1}. The mathematical equation modelling the spiral-shaped movement of manta rays in an n-d space can be defined by Eq. (9)-(11), in which, the term $\beta$ is defined as:
\begin{equation}
    \beta = 2 e^{r_1 \frac{T-t+1}{T}}\cdot sin(2\pi r_1)
\end{equation}
where, $\beta$ is a weight coefficient, $T$ is the max number of iterations, and $r_1$ is a rand number in the interval [0, 1]. In cyclone foraging, all individuals randomly perform the search with respect to the food source as their reference position accounting to the final exploitation for the region having the best solution found so far \cite{MRFO_1}. To enhance space exploration and to enable MRFO to achieve an extensive global search, the school of mantas is forced to search for a new position far from the current best one by assigning a new random in the entire search \cite{MRFO_1, MRFO_2}. To achieve this, the mathematical equation of a new random position is defined as follows:
\begin{equation}
    \textbf{x}_{rand} = \textbf{x}_l + (\textbf{x}_u - \textbf{x}_l) 
\end{equation}
$\textbf{x}_l = (\textbf{x}_l^1, \ldots, \textbf{x}_l^D)$, and $\textbf{x}_{rand} = (x_{rand}^1, \ldots, x_{rand}^d)$, such that, $ d= 1, \ldots, D$. Each $x_{rand}^d$ is the random position.  
\subsubsection{Somersault foraging}
In this behaviour, the position of plankton is viewed as a pivot, where each manta tends to swim to and fro around the pivot and somersault a new position. Each manta in the school updates its position around the best solution found so far whose mathematical model is defined by Eq. (10) as presented in \textbf{Algorithm 1}. $\textbf{S}$ denotes the somersault factor that determines the somersault range of manta rays $\textbf{S}=2$, $r_2$ and $r_3$ are two random numbers in $[0,~1]$.   
\subsection{Levy Flight}
The L\'{e}vy Flight trajectory was originally introduced by L\'{e}vy, and then Benoit Mandelbrot described it in detail. In general, L\'{e}vy flight is a type of random walk in which the steps are drawn from a L\'{e}vy distribution. A variety of studies have demonstrated that the behavior of flight for many animals and insects demonstrates the typical characteristics of  L\'{e}vy Flight. As described in \cite{LF_1}, some fruits flies or \textit{Drosophila melagonaster} search their landscape by using a set of straight flight paths puntuated by a sudden $90^o$ turn causing a L\'{e}vy Flight (LF) style intermittent scale-free search pattern. LF can be considered as a random walk in uncertain environments while delivering an efficient exploration and exploitation mechanism in unknown large search spaces \cite{LF_2}.  
\subsection{Proposed Evolutionary ELM using MRFO with LF}
This section introduces the proposed Evolutionary Extreme Learning Machine using Manta Ray Foraging Optimization with L\'{e}vy Flight (EELM-MRFO-LF). L\'{e}vy Flight is applied to improve exploration capabilities and to avoid getting trapped in a local optima while improving the convergence of traditional MRFO. 

\begin{algorithm*}

\DontPrintSemicolon
 Initialize MRFO parameters and each Manta Ray by: 
\begin{itemize}
    \item $\textbf{x}_i(t) = \textbf{x}_l$ + \textit{rand}$\cdot(\textbf{x}_u - \textbf{x}_l), i = 1, \ldots, N$
    \item Compute fitness of each individual $f_i = f(\textbf{x}_i)$ and obtain the best solution found so far $\textbf{x}_{best}$
\end{itemize}

\While{Stop criterion is not satisfied}{
  \ForAll{ $i = 1 $ to $N$ }
  {
    \eIf { $\textit{rand} < 0.5$ }
      {$// \textbf{Cyclone foraging}$
      
       \eIf { $t/T_{max}<\textit{rand}$ }
      {
       \begin{equation}
           \textbf{x}_{rand} = \textbf{x}_l + \textit{rand}\cdot(\textbf{x}_u - \textbf{x}_l)
       \end{equation}
      \begin{align}
            \textbf{x}_i(t+1)&=
          \begin{cases}
          \textbf{x}_{rand} + r\cdot(\textbf{x}_{rand} - \textbf{x}_i (t)) + \beta\cdot(\textbf{x}_{rand} - \textbf{x}_i (t));~i=1\\
          \textbf{x}_{rand} + r\cdot(\textbf{x}_{i-1}(t) - \textbf{x}_i (t)) + \beta\cdot(\textbf{x}_{rand} - \textbf{x}_i (t));~i=2, \ldots, N\\
         \end{cases}
      \end{align}
      }
      {
    \begin{align}
            \textbf{x}_i(t+1)&=
          \begin{cases}
          \textbf{x}_{best} + r\cdot(\textbf{x}_{best} - \textbf{x}_i (t)) + \beta\cdot(\textbf{x}_{best} - \textbf{x}_i (t));~i=1\\
          \textbf{x}_{best} + r\cdot(\textbf{x}_{i-1}(t) - \textbf{x}_i (t)) + \beta\cdot(\textbf{x}_{best} - \textbf{x}_i (t));~i=2, \ldots, N\\
         \end{cases}
      \end{align}
  }
      }
   {
   $// \textbf{Chain Foraging}$
     \begin{align}
            \textbf{x}_i(t+1)&=
          \begin{cases}
          \textbf{x}_{i}(t) + r\cdot(\textbf{x}_{best} - \textbf{x}_i (t)) + \beta\cdot(\textbf{x}_{i}(t) - \textbf{x}_i (t));~i=1\\
          \textbf{x}_{i-1}(t) + r\cdot(\textbf{x}_{i-1}(t) - \textbf{x}_i (t)) + \beta\cdot(\textbf{x}_{best} - \textbf{x}_i (t));~i=2, \ldots, N\\
         \end{cases}
      \end{align}
  }
  Compute the fitness of each individual $f(\textbf{x}_i(t+1))$
\uIf{ $f(\textbf{x}_i(t+1)) < f(\textbf{x}_{best})$ }
      {$\textbf{x}_{best} = \textbf{x}_i(t+1)$}
$//\textbf{Somersault foraging}$

  \ForAll{ $i = 1 $ to $N$ }
  {
   \begin{equation}
       \textbf{x}_i(t+1) = \textbf{x}_i(t)+ \textbf{S} \cdot  (r_2 \cdot \textbf{x}_{best} - r_3 \cdot \textbf{x}_i(t) ) 
   \end{equation}
  }
  Compute the fitness of each individual $f(\textbf{x}_i(t+1))$
\uIf{ $f(\textbf{x}_i(t+1)) < f(\textbf{x}_{best})$ }
      {$\textbf{x}_{best} = \textbf{x}_i(t+1)$}
  }
}
\caption{\small Manta Ray Foraging Optimization Algorithm}\label{PSO_algorithm}
\end{algorithm*}

This is reached by enhancing the diversity of agents during somersault foraging. Therefore, the position of each manta ray can be modelled during l\'{e}vy flight trajectory using the following formula:
\begin{equation}
    \textbf{x}(t+1) = \textbf{x}(t) + \mu sign[rand - 1/2]  \oplus Levy
\end{equation}
$\textbf{x}(t)$ indicates the $ith$ manta ray, $\mu$ is a random number that is consistent with a uniform distribution, the product $\oplus$ means entrywise multiplication, and rand is a random number in the interval $[0-1]$. Eq. (13) improves the ability of basic MRFO by leading the somersault foraging to randomly walk while removing local minima and ensuring that the algorithm can explore the search space efficiently. L\'{e}vy Flight offers a random walk based on the distribution $Levy \thicksim  u = t^{-\lambda}, 1 < \lambda \leq 3$.
L\'{e}vy Flight is a random walk in which the step length follows a distribution that is heavy-tailed. 

\begin{algorithm}
\DontPrintSemicolon
\KwIn{Training Data $\left(x_p,t_p\right)$}
\KwOut{Optimal weights $\hat{\textbf{W}}$ and $\hat{\pmb{\beta}}$ of selected ELM }
\SetKwBlock{Begin}{function}{end function}
\Begin($\text{EELM-MRFO-LF}$)
{  
  Define ELM parameters:  $\textbf{h}(\textbf{w}_k;\textbf{w}_p)$ and $M$ \;
  Initialize MRFO parameters:\:
  
  \hspace{4mm}\textbullet~$\textbf{w}_i(t) = \textbf{w}_l$ + \textit{rand}$\cdot(\textbf{w}_u - \textbf{w}_l), i = 1, \ldots, N$\:
  
  \hspace{4mm}\textbullet~For each manta ray, compute $\hat{\boldsymbol\beta}_{i,t=0} = \textbf{H}^{\dagger}_{t=0} \textbf{T}$\;
  
  \hspace{4mm}\textbullet~Compute the fitness of each manta ray $J_i(\textbf{w}_i)$\:
  
  \hspace{4mm}\textbullet~Find the best fitness $J_{i}(\textbf{w}_{best};\textbf{x}_p)$.
  
  \hspace{4mm}\textbullet~Set $t=1$\:
  
\While{$t \leq~Max_{T}$ }{
  \ForAll{ i = 1 to N }
  {
       \eIf { $\textit{rand} < 0.5~(Cyclone ~foraging)$ }
      {
       \eIf { $t/T_{max}<\textit{rand}$ }
      {
         $\textbf{w}_{rand} = \textbf{w}_l + \textit{rand}\cdot(\textbf{w}_u - \textbf{w}_l)\nonumber$
       Compute $\textbf{w}_i(t+1)$ using Eq. (6). 
      }
      {
     Compute $\textbf{w}_i(t+1)$ using Eq. (7).
  }
      }
   {
   $// Chain~Foraging$
    
       Compute $\textbf{w}_i(t+1)$ using Eq. (8).
  }
  Compute fitness: $J_i(\textbf{w}_i(t+1);\textbf{x}_p(t+1))$
  
\uIf{ $J_i(\textbf{w}_i(t+1);\textbf{x}_p(t+1)) < J_i(\textbf{w}_{best})$ }
      {$\textbf{w}_{best} = \textbf{w}_i(t+1)$}
  $//\textbf{Somersault foraging}$
  
 \ForAll{ $i = 1 $ to $N$ }
  {
     \begin{itemize}
      \item Compute $\textbf{w}_i(t+1)$ using Eq. (11)
   \item Update the position of current search \newline agent using L\'{e}vy Flight (Eq. (13))  \end{itemize}
  }
  Compute the fitness of each individual $J_i(\textbf{w}_i(t+1);\textbf{x}_p(t+1))$
\uIf{ $J_i(\textbf{w}_i(t+1);\textbf{x}_p(t+1)) < J_i(\textbf{w}_{best})$ }
      {$\textbf{w}_{best} = \textbf{w}_i(t+1)$}
  }
  $t = t+1$\;  
  }
  \label{endfor}
  \Return{($\textbf{w}_{best}$, $\hat{\beta}_{best}$) }
}
\caption{\small Proposed EELM-MRFO-LF algorithm}\label{PSO_algorithm}
\end{algorithm}

Mantegna's algorithm is employed to mimic a $\lambda-\text{stable}$ distribution by generating random step lengths $s$ that have the same behaviour of L\'{e}vy flights as $s =\mu/|v|^{1/\beta}$:
where $s$ is the step length of L\'{e}vy Flight, which is $levy(\lambda)$, and the term $\lambda = 1 + \beta$, where $\beta = 1.5$, $\mu = N(-, \sigma^2)$, and $v = N(-, \sigma^2)$ being both normal distributed with:
\begin{equation}
 \sigma_{\mu} = \left[ \frac{\Gamma (1+ \beta) \times  sin(\pi \times \beta/2)}{\Gamma ((1+\beta/2)\times \beta  \times 2^{(\beta-1)/2})} \right]^{1/\beta}
\end{equation}
The major contribution of implementing L\'{e}vy Flight is to enhance the diversification ability of MRFO by attempting to search around the best objective solution. The main steps of of the proposed EELM-MRFO-LF is described in Algorithm 2. According to Algorithm 2, once the ELM structure and the activation function $\textbf{h}(\textbf{w}_k; \textbf{x}_p)$ for each hidden unit is defined, a number of initial manta rays is initialised, and their corresponding fitness function is computed (See line 2-7). The position of each manta ray in the school is defined as follows: 
\begin{equation}
\textbf{w}_i = \left( \underbrace{ \overbrace{w_{11}, \ldots, w_{1M}}^{\text{\footnotesize Weight vector $\textbf{w}_1$}}, \ldots,  \overbrace{w_{1M}, \ldots, w_{nM}}^{\text{\footnotesize  Weight vector $\textbf{w}_n$}}}_{\text{\footnotesize Position of the $ith$ manta ray (agent)}} \right);~k=1, \ldots, n 
\end{equation}
RMSE is used to define the fitness of each agent as follows:
\begin{equation}
      J_i(\textbf{w}_i(t+1); \textbf{x}_p(t+1))= \left( \frac{1}{P} \sum_{p=1}^P (\textbf{h}_p(\textbf{w}_i;\textbf{x}_p) \hat{\boldsymbol\beta}_{i,t} - \textbf{t}_p)^2  \right)^{1/2}
\end{equation}
From line 5, the term $\hat{\boldsymbol\beta}_{i,t=0}$ denotes the optimal output weight for the $ith$ agent at $t=0$. Note when $J_i(t)$ is computed, the term $\hat{\boldsymbol\beta_i}$ is obtained using Eq. (5). $\textbf{w}_l$ and $\textbf{w}_u$ is the lower and upper search limit, respectively. At each iteration, cyclone foraging, chain foraging and somersault foraging is implemented (line 9-23). In line 23, each agent's position is finally updated with a random LF walk.

\section{Experiments and Discussions}
In this section, the performance of the proposed EELM-MRFO-LF is evaluated for the prediction of relaxed and unrelaxed formation energies to the ground state crystal structure. To compare the prediction accuracy of EELM-MRFO-LF with respect to state-of-art methodologies, techniques such as ELM optimized by continuous genetic algorithms (ELM-GA) \cite{ELM_AG}, ELM and Whale Optimization algorithm (ELM-WOA) \cite{ELM_WOA}, ELM and Particle swarm optimization (ELM-PSO), traditional MRFO and ELM (ELM-MRFO) and the basic version of ELM \cite{intro_2}  are implemented \cite{ARS1}. For cross-validation purposes, a 5-fold strategy was implemented for 20 random experiments, while the experiment setup for an EELM-MRFO-LF consists of two parameters. A SLFN with $300$ hidden nodes using a Sigmoid function is implemented for all ELM-based models. On the other hand, to implement MRFO-LF, the population size was set up to 20 manta rays, ($\textbf{x}_l, \textbf{x}_u) = [-1, 1]$ where the optimal number of training iterations was found equal to $50$. Two different metrics are computed to evaluate the prediction performance of each model, namely: a) RMSE and b) $R^2$ \cite{ARS1_11}. 

\begin{figurehere}
\begin{center}
\includegraphics[width=9.25cm, height=8.8cm]{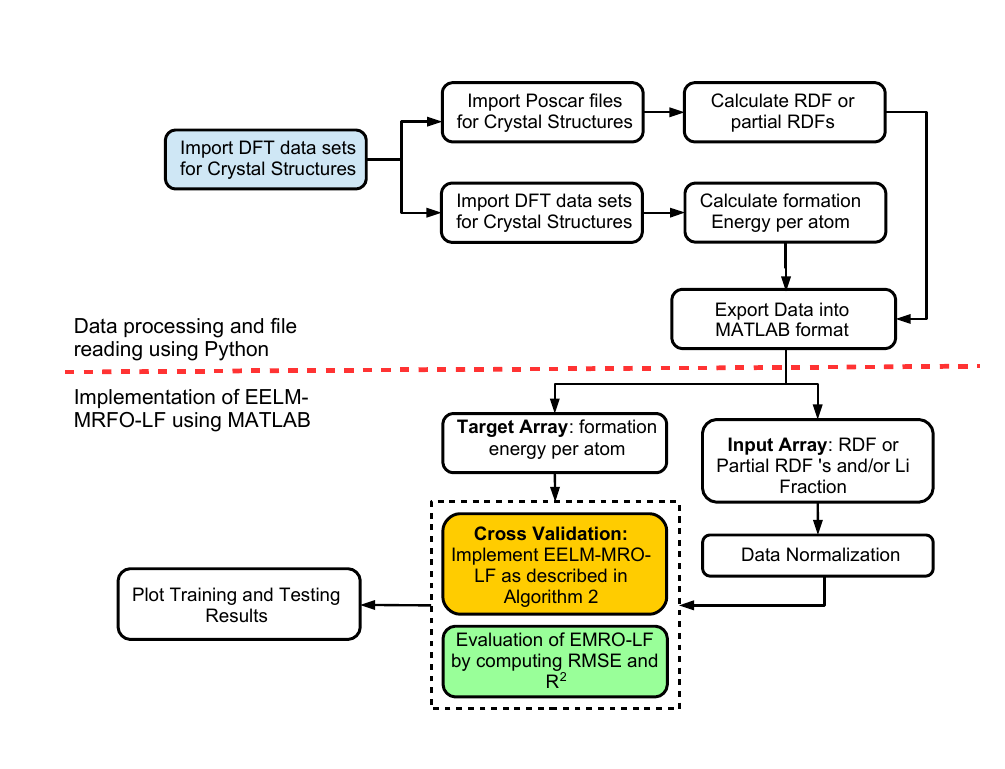}
\vspace{0mm}
\caption{Flow diagram of the implementation of EELM-MRFO-LF.}
\label{fig::image_preprocessing}
\end{center}
\end{figurehere}

\begin{tablehere}
\caption{ Comparison of Average performance for twenty random runs using RMSE in $\text{meV/Atoms}$ and data fit ($R^2$).}\label{iris_results} 
\centering 
\begin{tabular}{p{3.0cm} |p{1cm}|p{1.5cm} |p{0.9cm}|p{1cm} |p{1cm}| p{1cm} | p{1.0cm} }
\hline 
 \scriptsize Models &\multicolumn{3}{c}{\scriptsize Training ($\%$)}  & \multicolumn{4}{c}{\scriptsize Testing ($\%$)} \\
 
\hline
& \multicolumn{2}{c}{\scriptsize $R^2$}&\scriptsize Time (s)&\multicolumn{3}{c}{\scriptsize RMSE}&\multicolumn{1}{c}{ \scriptsize $R^2$ }\\   

\hline 
  \scriptsize ELM-GA-unrelaxed & \multicolumn{2}{c}{\scriptsize 0.991} & \multicolumn{1}{c}{\scriptsize 1180.1} &\multicolumn{3}{c}{\scriptsize 11.31} & \multicolumn{1}{c}{\scriptsize 0.990}\\
  \scriptsize ELM-GA-relaxed & \multicolumn{2}{c}{\scriptsize 0.995} & \multicolumn{1}{c}{\scriptsize 1150.3} &\multicolumn{3}{c}{\scriptsize 10.42} & \multicolumn{1}{c}{\scriptsize 0.994 }\\
  \scriptsize ELM-WOA-unrelaxed & \multicolumn{2}{c}{\scriptsize 0.992} & \multicolumn{1}{c}{\scriptsize 1233.1} &\multicolumn{3}{c}{\scriptsize 11.09} & \multicolumn{1}{c}{\scriptsize 0.993 }\\
  \scriptsize ELM-WOA-relaxed & \multicolumn{2}{c}{\scriptsize 0.994} & \multicolumn{1}{c}{\scriptsize 1158.8} &\multicolumn{3}{c}{\scriptsize 10.81} & \multicolumn{1}{c}{\scriptsize 0.993 }\\
 \scriptsize  ELM-PSO-unrelaxed & \multicolumn{2}{c}{\scriptsize 0.990} & \multicolumn{1}{c}{\scriptsize 1055.5} &\multicolumn{3}{c}{\scriptsize 9.800} & \multicolumn{1}{c}{\scriptsize  0.989}\\
 \scriptsize  ELM-PSO-relaxed & \multicolumn{2}{c}{\scriptsize 0.996} & \multicolumn{1}{c}{\scriptsize 1042.5} &\multicolumn{3}{c}{\scriptsize 9.350} & \multicolumn{1}{c}{\scriptsize  0.994}\\
  \scriptsize  EELM-MRFO-LF-unrelaxed& \multicolumn{2}{c}{\scriptsize 0.996} & \multicolumn{1}{c}{\scriptsize 1222.3} &\multicolumn{3}{c}{\scriptsize 9.300}&\multicolumn{1}{c}{\scriptsize \textbf{ 0.995} }\\
   \scriptsize  EELM-MRFO-LF-relaxed& \multicolumn{2}{c}{\scriptsize 0.997} & \multicolumn{1}{c}{\scriptsize 1198.1} &\multicolumn{3}{c}{\scriptsize 9.100}&\multicolumn{1}{c}{\scriptsize  \textbf{0.996}}\\
  \scriptsize ELM-unrelaxed & \multicolumn{2}{c}{\scriptsize 0.977} & \multicolumn{1}{c}{\scriptsize 25.01} &\multicolumn{3}{c}{\scriptsize 19.97}&\multicolumn{1}{c}{\scriptsize  0.957}\\
  \scriptsize ELM-relaxed & \multicolumn{2}{c}{\scriptsize 0.989} & \multicolumn{1}{c}{\scriptsize 24.22} &\multicolumn{3}{c}{\scriptsize 19.80}&\multicolumn{1}{c}{\scriptsize  0.987}\\
  \scriptsize EELM-MRFO-unrelaxed & \multicolumn{2}{c}{\scriptsize 0.994} & \multicolumn{1}{c}{\scriptsize 1176.9} &\multicolumn{3}{c}{\scriptsize 10.52}&\multicolumn{1}{c}{\scriptsize  0.991}\\
  \scriptsize EELM-MRFO-relaxed & \multicolumn{2}{c}{\scriptsize 0.996} & \multicolumn{1}{c}{\scriptsize 1108.2} &\multicolumn{3}{c}{\scriptsize 10.39}&\multicolumn{1}{c}{\scriptsize  0.993}\\
\hline
		\end{tabular}
		\centering 
		\label{nfm_results} 
\end{tablehere}
\vspace{4mm}

To perform all experiments, the workflow presented in Fig. 1 is used to implement the proposed EELM-MRFO-LF using MATLAB and Python. Similar to \cite{b2}, we calculate the average formation energy prediction errors of 20 random experiments over different partitions of training and testing.

\begin{figurehere}
\begin{center}
\includegraphics[width=8.8cm, height=5.2cm]{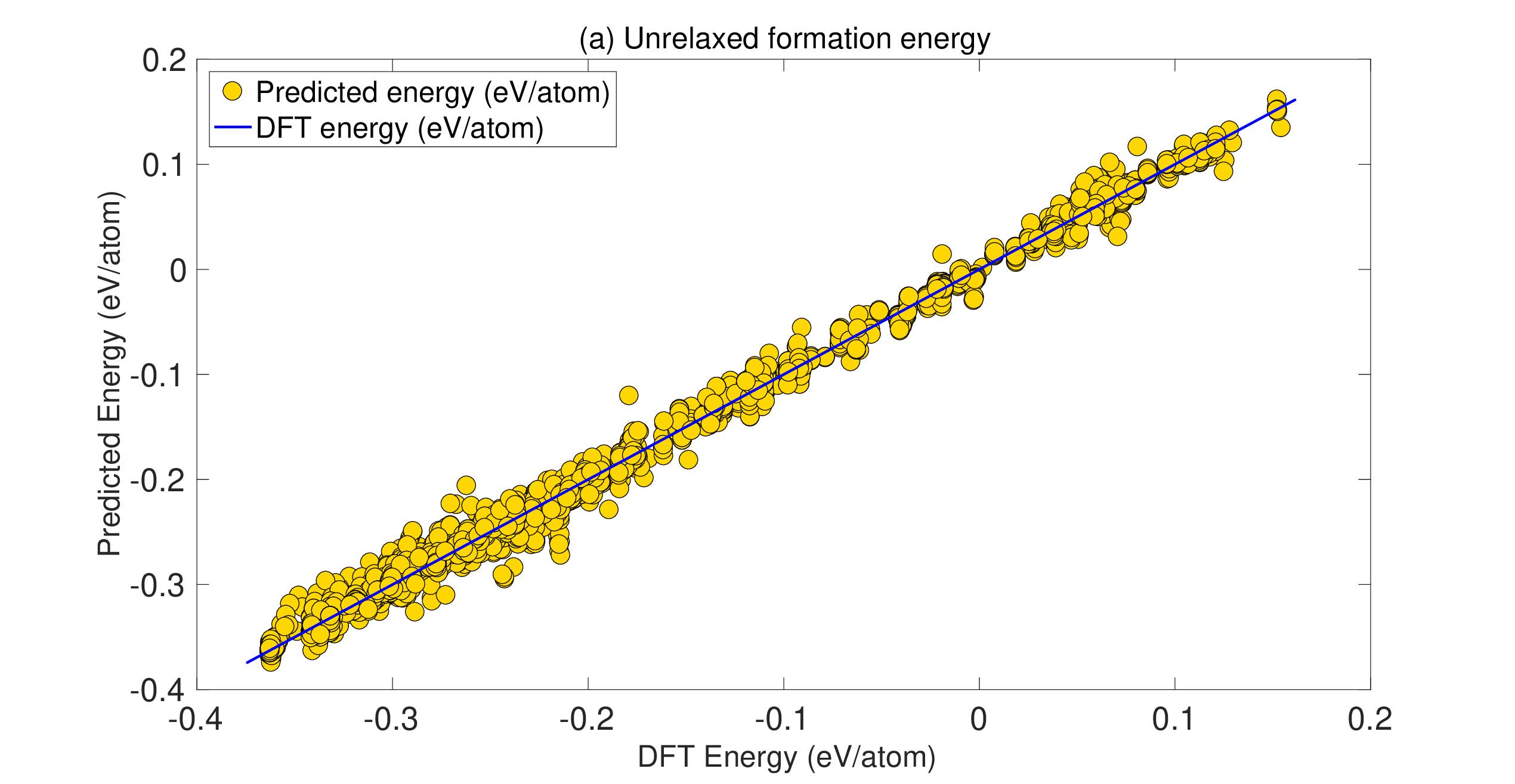}
\includegraphics[width=8.8cm, height=5.4cm]{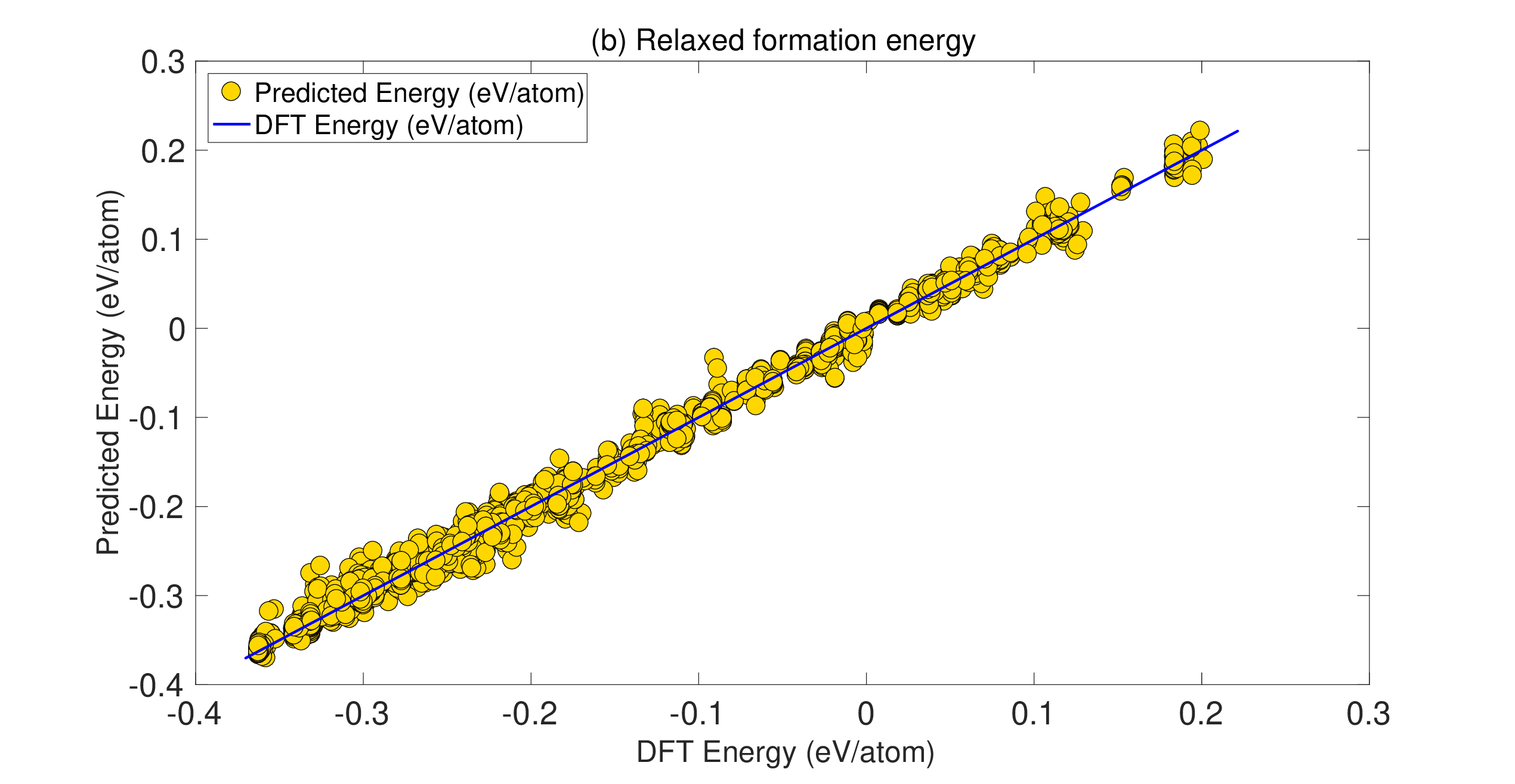}
\vspace{0mm}
\caption{EELM-MRFO-LF-predicted vs DFT-calculated (a) unrelaxed and (b) relaxed formation energies of structures in the testing set.}
\label{fig::image_preprocessing}
\end{center}
\end{figurehere}

\begin{figurehere}
\begin{center}
\hspace{1mm}
\includegraphics[width=8.3cm, height=5.0cm]{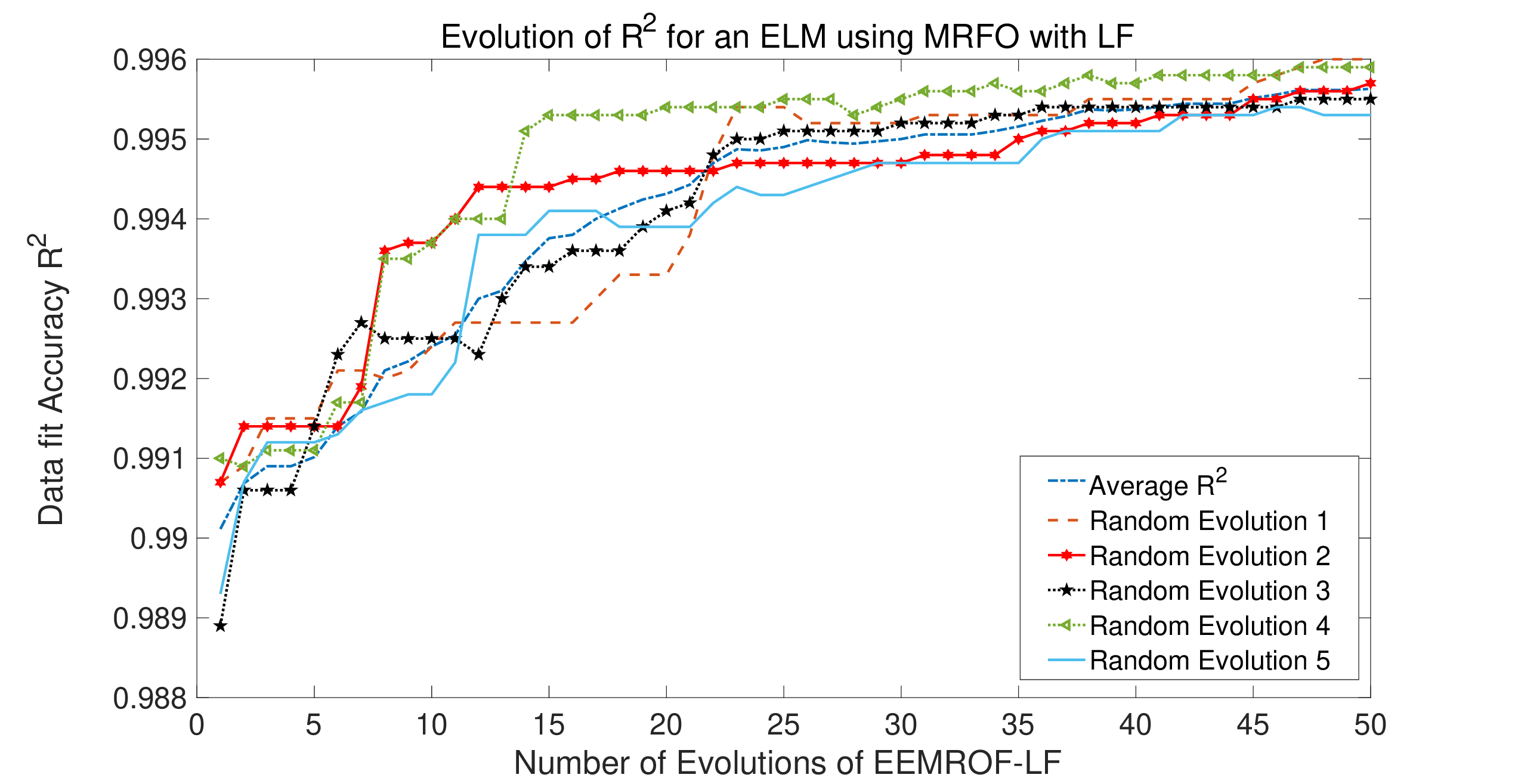}
\vspace{0mm}
\caption{Average performance of twenty runs of an EELM-MRFO-LF-unrelaxed with testing data and its comparison with five random experiments.}
\label{fig::image_preprocessing}
\end{center}
\end{figurehere}
\begin{figurehere}
\begin{center}
\hspace{1mm}
\includegraphics[width=8.4cm, height=5.0cm]{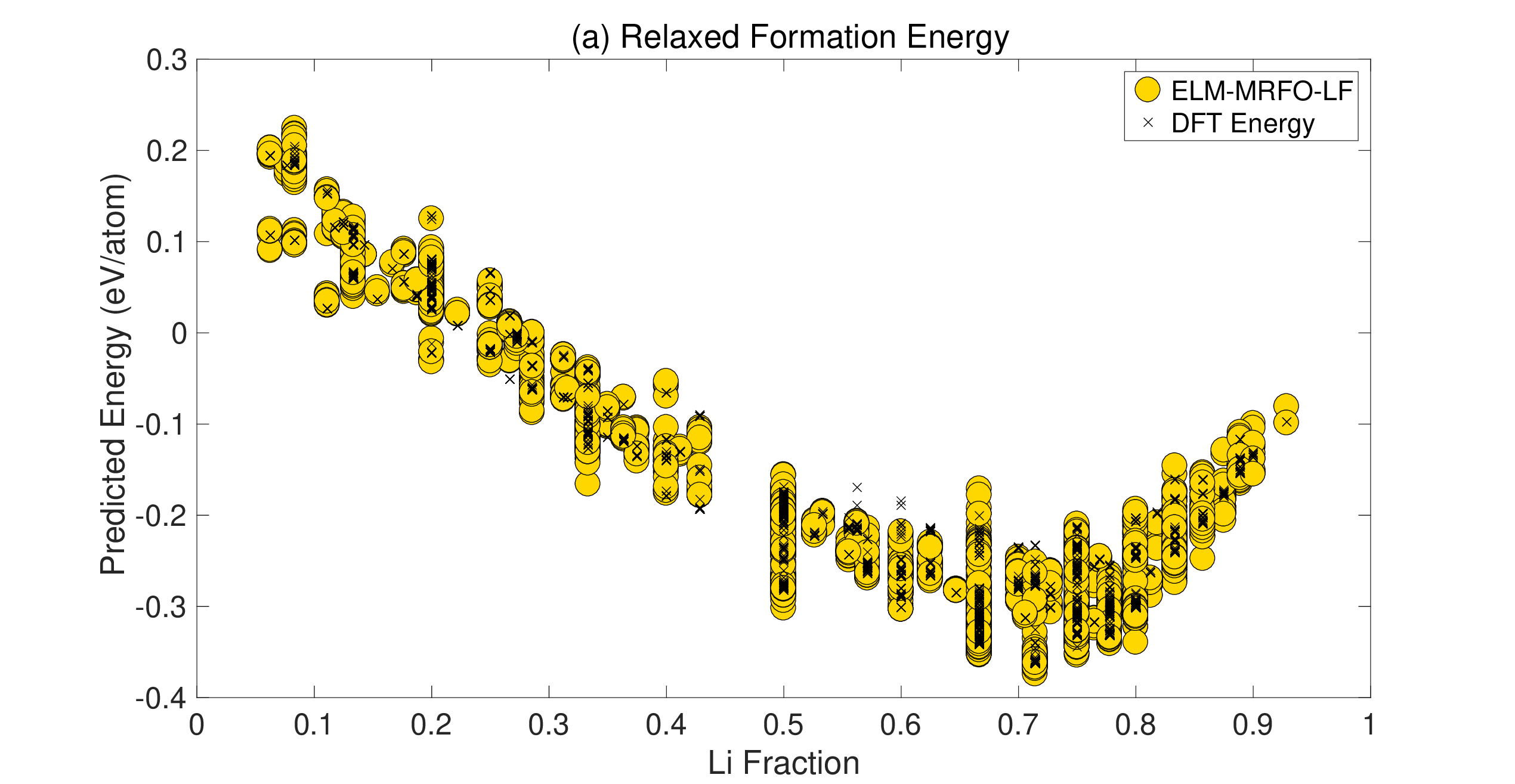}

\includegraphics[width=8.4cm, height=5.0cm]{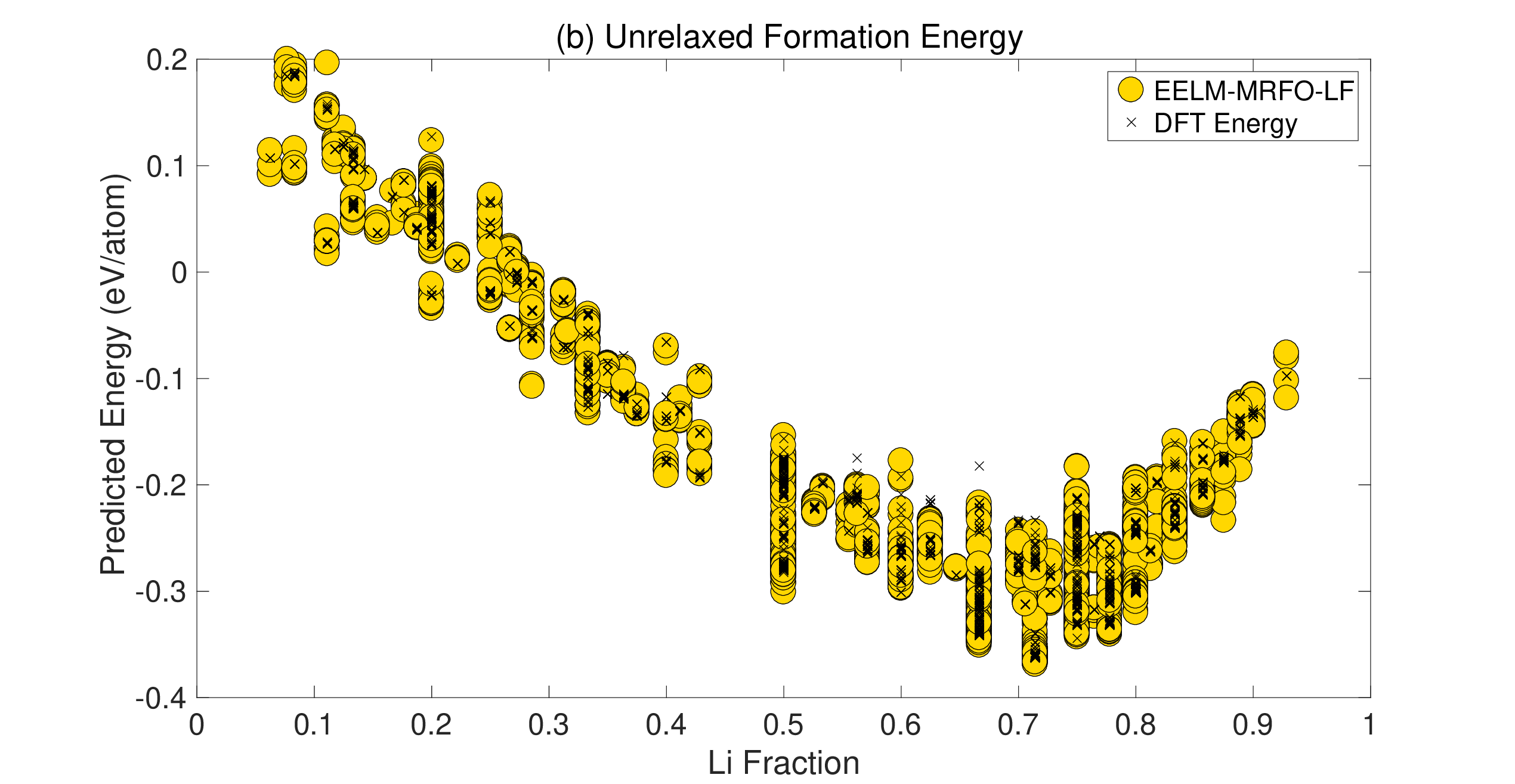}

\caption{Phase diagrams of the Li-Ge system showing (a) unrelaxed and (b) relaxed formation energies obtained from EELM-MRFO vs. DFT-calculated formation energies of structures in the testing set.}
\label{fig::image_preprocessing}
\end{center}
\end{figurehere}
\vspace{3mm}

In Table I, the average RMSE and $R^2$ for relaxed - unrelaxed energies produced by the proposed EELM-MRFO is compared with other techniques. As described in \cite{b2}, predicting relaxed formation energies - formation energies of the minimum-energy configurations makes possible to predict whether a new structure generated using an evolutionary technique will relax to a low-energy configuration or not \cite{b2}.  As described in Table I, the proposed EELM-MRFO-LF provides a good balance between training and testing performance where the average RMSE for predicting unrelaxed energies is 9.3 me/V and 10.52 me/V for traditional EELM-MRFO. From Table I, it can also be observed that using together MRFO and L\'{e}vy Flight improves not only data fit capabilities of ELM but also its generalisation properties. Fig. 2, compares the results of a random experiment for the predicted unrelaxed vs relaxed formation energies in the Li-Ge system with the DFT data for the proposed EELM-MRFO-LF. In Fig. 3, to illustrate the evolution of  EELM-MRFO-LF for testing data, its average data fit accuracy is compared to the evolution of five random experiments. Finally, Fig. 4 shows unrelaxed vs relaxed formation energies in Li-Ge system with the DFT data for the proposed EELM-MRFO-LF. Similar to the results presented in Fig. 2, in Fig. 4, the predicted formation energies show a lack of any outliers for the prediction of unseen data in the composition space.






\section*{CONCLUSIONS}
In this paper, an Evolutionary Extreme Learning Machine using Manta Ray Foraging Optimization with L\'{e}vy Flight (EELM-MRFO-LF) for the prediction of unrelaxed and relaxed formation energies of compounds relative to the ground state crystal structure is presented. The proposed strategy is a machine learning algorithm for the training of Single Layer Feedforward Neural Networks (SLFNs). Compared to traditional MRFO, an EELM-MRFO-LF provides a superior testing via enhancing the diversification ability of traditional MRFO to search around the best objective function. An EELM-MRFO-LF iteratively improves exploration capabilities and avoid getting trapped in a local optima while improving the convergence properties of an MRFO. 

Future work will focus on implementing the proposed strategy in solving other practical engineering optimization problems in the field of materials science as well as new hybrid meta-hybrid version of MRFO.

\end{document}